\documentclass[10pt,twocolumn,letterpaper]{article}

\usepackage{cvpr}
\usepackage{times}
\usepackage{epsfig}
\usepackage{graphicx}
\usepackage{amsmath}
\usepackage{amssymb}

\usepackage[pagebackref=true,breaklinks=true,letterpaper=true,colorlinks,bookmarks=false]{hyperref}

\usepackage{float}
\usepackage{multirow}
\usepackage{array}
\usepackage{booktabs}
\usepackage{marvosym}
\usepackage{fontawesome}
\usepackage{lipsum}

\newcommand\blfootnote[1]{%
  \begingroup
  \renewcommand\thefootnote{}\footnote{#1}%
  \addtocounter{footnote}{-1}%
  \endgroup
}

\newcommand{\parahead}[1]{\par\textbf{#1}:\ }

\newenvironment{packed_itemize}
{\begin{itemize}
    \setlength{\itemsep}{1pt}
    \setlength{\parskip}{0pt}
    \setlength{\parsep}{0pt}
}{\end{itemize}}

\newcommand{\filluptopage}[1]{%
  \clearpage
  \loop\ifnum\value{page}<#1\relax
    \null\clearpage
  \repeat
  \loop\ifnum\value{page}=#1\relax
    \null\clearpage
  \repeat
}

\cvprfinalcopy 
\ifcvprfinal\fi
\begin{document}
\title{Normalized Object Coordinate Space for Category-Level\\ 6D Object Pose and Size Estimation}

\author{He Wang\textsuperscript{1} \quad Srinath Sridhar\textsuperscript{1} \quad Jingwei Huang\textsuperscript{1} \quad Julien Valentin\textsuperscript{2}\\ \quad Shuran Song\textsuperscript{3} \quad Leonidas J.~Guibas\textsuperscript{1,4}\\
\textsuperscript{1}Stanford University \quad  \textsuperscript{2}Google Inc. \quad \textsuperscript{3}Princeton University \quad \textsuperscript{4}Facebook AI Research}

\maketitle

\begin{abstract}
  The goal of this paper is to estimate the 6D pose and dimensions of unseen object instances in an RGB-D image.
  Contrary to ``instance-level'' 6D pose estimation tasks, our problem assumes that no exact object CAD models are available during either training or testing time.
  To handle different and unseen object instances in a given category, we introduce \textbf{Normalized Object Coordinate Space (NOCS)}---a shared canonical representation for all possible object instances within a category.
  Our region-based neural network is then trained to directly infer the correspondence from observed pixels to this shared object representation (NOCS) along with other object information such as class label and instance mask.
  These predictions can be combined with the depth map to jointly estimate the metric 6D pose and dimensions of multiple objects in a cluttered scene.
  To train our network, we present a new context-aware technique to generate large amounts of fully annotated mixed reality data.
  To further improve our model and evaluate its performance on real data, we also provide a fully annotated real-world dataset with large environment and instance variation.
  Extensive experiments demonstrate that the proposed method is able to robustly estimate the pose and size of unseen object instances in real environments while also achieving state-of-the-art performance on standard 6D pose estimation benchmarks. 
\end{abstract}

\blfootnote{\faGlobe~\url{https://hughw19.github.io/NOCS_CVPR2019}}
\section{Introduction}
\label{sec:intro}
Detecting objects, and estimating their 3D position, orientation and size is an important requirement in virtual and augmented reality (AR), robotics, and 3D scene understanding. These applications require operation in new environments that may contain previously unseen object instances. Past work has explored the \emph{instance-level 6D pose estimation} problem~\cite{rad2017bb8,tekin2017,kehl2017ssd,xiang2017posecnn,brachmann2016uncertainty,kouskouridas2016latent} where exact CAD models and their sizes are available beforehand. Unfortunately, these techniques cannot be used in general settings where the vast majority of the objects have never been seen before and have no known CAD models. On the other hand, \emph{category-level 3D object detection} methods~\cite{song2016deep,qi2017frustum,chen2016monocular,mousavian20173d,xiang2015data,deng2017amodal} can estimate object class labels and 3D bounding boxes without requiring exact CAD models. However, the estimated 3D bounding boxes are viewpoint-dependent and do not encode the precise orientation of objects. Thus, both these classes of methods fall short of the requirements of applications that need the 6D pose and 3 non-uniform scale parameters (encoding dimensions) of unseen objects.
\begin{figure}[t]
\centering
  \includegraphics[width=\columnwidth]{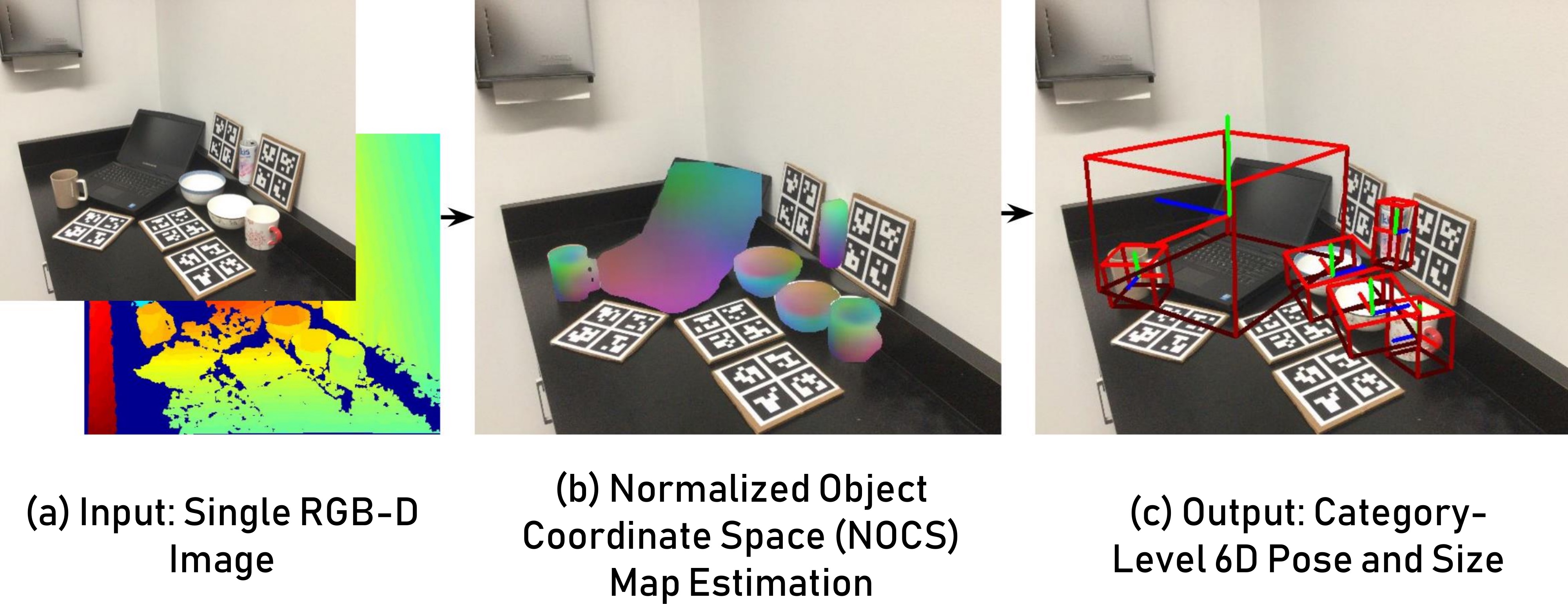}
  \caption{We present a method for category-level 6D pose and size estimation of multiple unseen objects in an RGB-D image. A novel normalized object coordinate space (NOCS) representation (color-coded in (b)) allows us to consistently define 6D pose at the category-level. We obtain the full metric 6D pose (axes in (c)) and the dimensions (red bounding boxes in (c)) for unseen objects.}
  \label{fig:teaser}
  \vspace{-3mm}
\end{figure}

In this paper, we aim to bridge the gap between these two families of approaches by presenting, to our knowledge, the first method for \textbf{category-level 6D pose and size estimation} of multiple objects---a challenging problem for novel object instances.
Since we cannot use CAD models for unseen objects, the first challenge is to find a representation that allows definition of 6D pose and size for different objects in a particular category. The second challenge is the unavailability of large-scale datasets for training and testing. Datasets such as SUN RGB-D~\cite{song2015sun} or NYU v2~\cite{Silberman:ECCV12} lack annotations for precise 6D pose and size, or do not contain table-scale object categories---exactly the types of objects that arise in table-top or desktop manipulation tasks for which knowing the 6D pose and size would be useful.

To address the representation challenge, we formulate the problem as finding correspondences between object pixels to normalized coordinates in a shared object description space (see Section~\ref{sec:overview}). We define a shared space called the \textbf{Normalized Object Coordinate Space} (NOCS) in which all objects are contained within a common normalized space, and all instances within a category are consistently oriented. This enables 6D pose and size estimation, even for unseen object instances. At the core of our method is a convolutional neural network (CNN) that jointly estimates the object class, instance mask, and a \emph{NOCS map} of multiple objects from a single RGB image. Intuitively, the NOCS map captures the normalized shape of the visible parts of the object by predicting dense correspondences between object pixels and the NOCS. Our CNN estimates the NOCS map by formulating it either as a pixel regression or classification problem. The NOCS map is then used with the depth map to estimate the full metric 6D pose and size of the objects using a pose fitting method.

To address the data challenge, we introduce a spatially context-aware mixed reality method to automatically generate large amounts of data (275K training, 25K testing) composed of realistic-looking synthetic objects from ShapeNetCore~\cite{chang2015shapenet} composited with real tabletop scenes. This approach allows the automatic generation of realistic data with object clutter and full ground truth annotations for class label, instance mask, NOCS map, 6D pose, and size. We also present \textbf{a real-world dataset for training and testing} with 18 different scenes and ground truth 6D pose and size annotations for 6 object categories, and in total 42 unique instances. To our knowledge, ours is the largest and most comprehensive training and testing datasets for 6D pose and size, and 3D object detection tasks.

Our method uses input from a commodity RGB-D sensor and is designed to handle both symmetric and asymmetric objects, making it suitable for many applications. Figure~\ref{fig:teaser} shows examples of our method operating on a tabletop scene with multiple objects unseen during training. In summary, the main contributions of this work are:
\begin{packed_itemize}
\item Normalized Object Coordinate Space (NOCS), a unified shared space that allows different but related objects to have a common reference frame enabling 6D pose and size estimation of unseen objects.
\item A CNN that jointly predicts class label, instance mask, and NOCS map of multiple unseen objects in RGB images. We use the NOCS map together with the depth map in a pose fitting algorithm to estimate the full metric 6D pose and dimensions of objects.
\item \textbf{Datasets}: A spatially context-aware mixed reality technique to composite synthetic objects within real images allowing us to generate a large annotated dataset to train our CNN. We also present fully annotated real-world datasets for training and testing.
\end{packed_itemize}

\section{Related Work}
\label{sec:relwork}
In this section, we focus on reviewing related work on category-level 3D object detection, instance-level 6D pose estimation, category-level 4~DoF pose estimation from RGB-D images, and different data generation strategies. 

\parahead{Category-Level 3D Object Detection}
One of the challenges in predicting the 6D pose and size of objects is localizing them in the scene and finding their physical sizes, which can be formulated as a 3D detection problem~\cite{Zhang2017DeepContext,depthRCNN,guptaCVPR13,RGBDroom,engelcke2017vote3deep}. Notable attempts include \cite{song2016deep,zhou2017voxelnet} who take 3D volumetric data as input to directly detect objects in 3D. Another line of work~\cite{qi2017frustum,gupta2015aligning,chen2017multi,lahoud20172d} proposes to first produce 2D object proposals in 2D image and then project the proposal into 3D space to further refine the final 3D bounding box location. The techniques described above reach impressive 3D detection rates, but unfortunately solely focus on finding the bounding volume of objects and do not predict the 6D pose of the objects.

\parahead{Instance-Level 6~DoF Pose Estimation}
Given its practical importance, there is a large body of work focusing on instance-level 6D pose estimation. Here, the task is to infer the 3D location and 3D rotation of objects (no scale), assuming exact 3D CAD models and size of these objects are available during training. The state of the art can be broadly categorized as template matching or object coordinates regression techniques. Template matching techniques align 3D CAD models to observed 3D point clouds with algorithms such as iterative closest point~\cite{besl1992method,zeng2017multi}, or use hand crafted local descriptors to further guide the alignment process \cite{LINEMOD,collet2011moped}. This family of techniques often suffer from inter- and intra-object occlusions, which is typical when we have only partial scans of objects. The second category of approaches based on object coordinates regression aim to regress the object surface position corresponding to each object pixel. Such techniques have been successfully employed for body pose estimation~\cite{taylor2012vitruvian,guler2018densepose}, camera relocalization~\cite{shotton2013scene,valentin2015exploiting} and 6D object pose estimation~\cite{brachmann2014learning}.

Both the above approaches need \emph{exact 3D models} of the objects during training and test time. Besides the practical limitation in storing all 3D CAD models or learned object coordinate regressors in memory at test time, capturing high-fidelity and complete 3D models of a very large array of objects is a challenging task. Although our approach is inspired by object coordinate regression techniques, it also significantly differs from the above approaches since we no longer require complete and high-fidelity 3D CAD models of objects at test time.

\parahead{Category-Level 4~DoF Pose Estimation}
There has been some work on category-level pose estimation~\cite{gupta2015aligning,SlidingShapes,guo2014scene,papon2015semantic,braun2016pose}, however they all make simplifying assumptions. First, these algorithms constrain the rotation prediction to be only along the gravity direction (only four degrees of freedom). Second, they focus on a few big room-scale object categories (\eg, chairs, sofa, beds or cars) and do not take object symmetry into account~\cite{gupta2015aligning,SlidingShapes,guo2014scene}. On the contrary, we estimate the pose of a variety of hand-scale objects, which are often much more challenging than the bigger room-scale objects due to larger pose variation. Our method also predicts full 6D pose and size without assuming the object’s gravity direction. Finally, our method runs at interactive frame rates (0.5~s per frame), which is significantly faster than alternative approaches ($\thicksim$70~s per frame for \cite{gupta2015aligning}, 25 mins per frame for \cite{SlidingShapes}).

\parahead{Training Data Generation} 
A major challenge with training CNNs is the lack of training data with sufficient category, instance, pose, clutter, and lighting variation. There have been several efforts aimed at constructing real-world datasets containing object labels (\eg, \cite{Silberman:ECCV12,song2015sun,xiang2014beyond}). Unfortunately, these datasets tend to be relatively small, mostly due to the high cost (time and money) associated with ground truth annotation. This limitation is a motivator for other works (\eg, \cite{papon2015semantic,song2017semantic,xiang2017posecnn}) which generate data that is exclusively synthetic allowing the generation of large amounts of perfectly annotated training data at a smaller cost. For the sake of simplicity, all these datasets ignore a combination of factors (material, sensor noise, and lighting) which creates a de-facto domain gap between the synthetic and real data distributions. To reduce this gap, \cite{dosovitskiy2015flownet} have generated datasets that mix real and synthetic data by rendering virtual objects on real backgrounds. While the backgrounds are realistic, the rendered objects are flying mid-air and out of context \cite{dosovitskiy2015flownet}, which prevent algorithms from making use of important contextual cues.

We introduce a new mixed reality method to automatically generate large amounts of data composed of synthetic renderings of objects and real backgrounds in a context-aware manner which makes it more realistic.
This is supported by experiments that show that our context-aware training data enables the model to generalize better to real-word test data. We also present a real-world dataset to further improve learning and for evaluation.

\begin{figure}[th!]
 \vspace{-3mm}
\centering
  \includegraphics[width=\columnwidth]{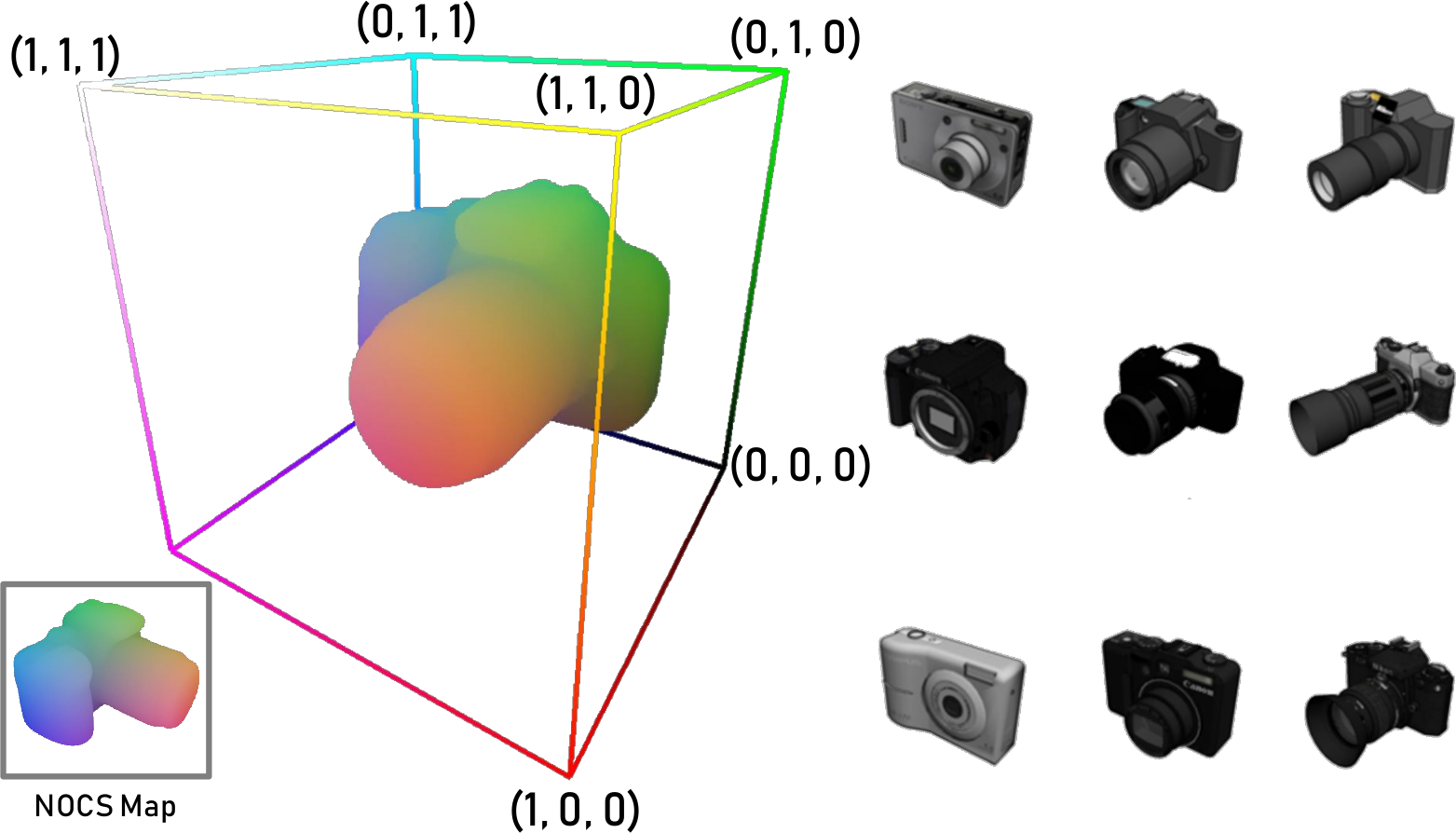}
  \caption{The Normalized Object Coordinate Space (NOCS) is a 3D space contained within a unit cube. For a given object category, we use canonically oriented instances and normalize them to lie within the NOCS. Each $(x, y, z)$ position in the NOCS is visualized as an RGB color tuple. We train our network on the perspective projection of the NOCS on the RGB image, the NOCS map (bottom left inset). At test time, the network regresses the NOCS map which is then used together with the depth map for 6D pose and size estimation.}
  \label{fig:nocs}
   \vspace{-3mm}
\end{figure}
\begin{figure*}[th!]
\centering
 \vspace{-7mm}
  \includegraphics[width=\linewidth]{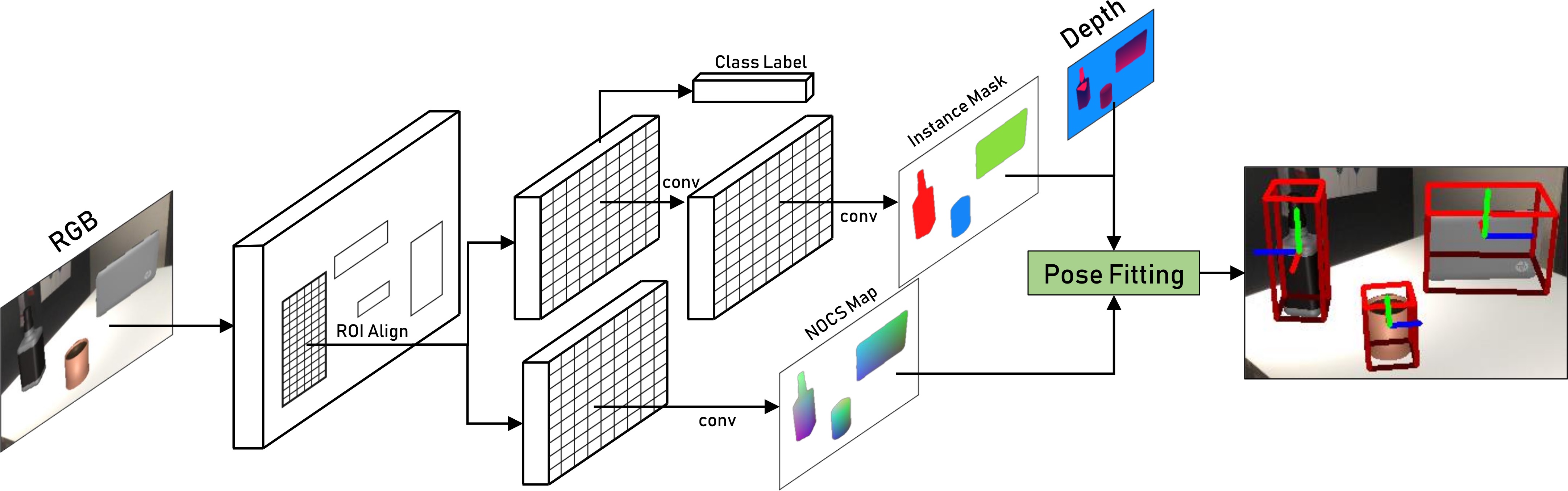} \vspace{-4mm}
  \caption{The inputs to our method are the RGB and depth images of a scene with multiple objects. Our CNN predicts the class label, instance mask, and NOCS map (color-coded) for each object in the RGB image. We then use the NOCS maps for each object together with the depth image to obtain the full metric 6D pose and size (axes and tight red bounding boxes), even if the object was never seen before.}
  \label{fig:pipeline}
  \vspace{-3mm}
\end{figure*}
\section{Background and Overview}
\label{sec:overview}

\parahead{Category-Level 6D Object Pose and Size Estimation}
We focus on the problem of estimating the 3 rotation, 3 translation, and 3 scale parameters (dimensions) of object instances. The solution to this problem can be visualized as a tight oriented bounding box around an object (see Figure~\ref{fig:teaser}). Although not previously observed, these objects come from known object categories (\eg,~\emph{camera}) for which training samples have been observed during training. This task is particularly challenging since we cannot use CAD models at test time and 6D pose is not well-defined for unseen objects. To overcome this, we propose a new representation that defines a shared object space enabling the definition of 6D pose and size for unseen objects.

\parahead{Normalized Object Coordinate Space (NOCS)}
The NOCS is defined as a 3D space contained within a unit cube \ie,~$\{x, y, z\} \in [0, 1]$. Given a shape collection of known object CAD models for each category, we normalize their size by uniformly scaling the object such that the diagonal of its tight bounding box has a length of 1 and is centered within the NOCS space (see Figure~\ref{fig:nocs}). Furthermore, we align the object center and orientation consistently across the same category. We use models from ShapeNetCore~\cite{chang2015shapenet} which are already canonicalized for scale, position, and orientation. Figure~\ref{fig:nocs} shows examples of canonicalized shapes in the \emph{camera} category. Our representation allows each vertex of a shape to be represented as a tuple $(x, y, z)$ within the NOCS (color coded in Figure~\ref{fig:nocs}).

Our CNN predicts the 2D perspective projection of the color-coded NOCS coordinates, \ie, a \emph{NOCS map} (bottom left in Figure~\ref{fig:nocs}). There are multiple ways to interpret a NOCS map: (1)~as a \textbf{shape reconstruction} in NOCS of the observed parts of the object, or (2)~as \textbf{dense pixel--NOCS correspondences}. Our CNN learns to generalize shape prediction for unseen objects, or alternatively learns to predict object pixel--NOCS correspondences when trained on a large shape collection. This representation is more robust than other approaches (\eg, bounding boxes) since we can operate even when the object is only partially visible.

\parahead{Method Overview}
Figure~\ref{fig:pipeline} illustrates our approach which uses an RGB image and a depth map as input. The CNN estimates the class label, instance mask, and the NOCS map from only the RGB image. We do not use the depth map in the CNN because we would like to exploit existing RGB datasets like COCO, which do not contain depth, to improve performance. The NOCS map encodes the shape and size of the objects in a normalized space. We can therefore use the depth map at a later stage to lift this normalized space, and to predict the full metric 6D object pose and size using robust outlier removal and alignment techniques.

Our CNN is built upon the Mask R-CNN framework~\cite{he2017mask} with improvements to jointly predict NOCS maps in addition to class labels, and instance masks. Section~\ref{sec:method} contains more details on our improvements and new loss functions that can handle symmetric objects. During training, we use ground truth images rendered with a new Context-Aware MixEd ReAlity (CAMERA) approach (see Section~\ref{sec:dataset}). This large dataset allows us to generalize to new instances from new categories at testing time. To further bridge the domain gap we also use a smaller real-world dataset.

\begin{figure*}[th!]
\centering
\vspace{-5mm}
  \includegraphics[width=\textwidth]{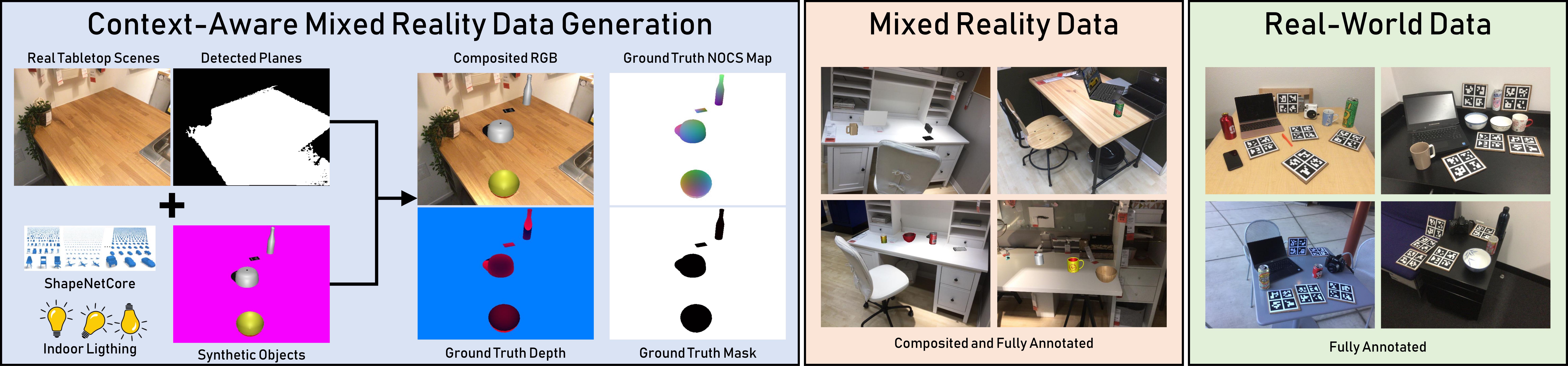}
  \vspace{-2mm}
  \caption{We use a Context-Aware MixEd ReAlity (CAMERA) approach to generate data by combining real images of tabletop scenes, detect planar surfaces, and render synthetic objects onto the planar surfaces (left). Since the objects are synthetic, we obtain accurate ground truth for class label, instance mask, NOCS map, and 6D pose and size. Our approach is fast, cost-effective, and results in realistic and plausible images (middle). We also gather a real-world dataset for training, testing, and validation (right).}
  \label{fig:dataset}
  \vspace{-3mm}
\end{figure*}
\section{Datasets}
\label{sec:dataset}

A major challenge in category-level 3D detection, and 6D pose and size estimation is the unavailability of ground truth data. While there have been several attempts like NYU v2~\cite{Silberman:ECCV12} and SUNRGB-D~\cite{song2015sun}, they have important limitations. First, they do not provide 6D pose of objects and focus on just 3D bounding boxes. Second, applications such as augmented reality and robotics benefit from hand-scale objects in tabletop settings which are missing from current datasets which focus on on larger objects such as as chairs and tables. Finally, these datasets do not contain annotations for the type of ground truth we need (\ie,~NOCS maps) and contain limited number of examples.

\subsection{Context-Aware Mixed Reality Approach}
\label{sec:camera_dataset}
To facilitate the generation of large amounts of training data with ground truth for hand-scale objects, we propose a new \textbf{Context-Aware MixEd ReAlity (CAMERA)} approach which addresses the limitations of previous approaches, and makes data generation less time consuming and significantly more cost-effective. It combines real background images with synthetically rendered foreground objects in a \emph{context-aware} manner \ie, the synthetic objects are rendered and composited into real scenes with plausible physical locations, lighting, and scale (see Figure~\ref{fig:dataset}). This \emph{mixed reality} approach allows us to generate significantly larger amounts of training data than previously available.

\parahead{Real Scenes}
We use real RGB-D images of 31~widely vaying indoor scenes as background (Figure~\ref{fig:dataset} middle). Our focus is on tabletop scenes since the majority of indoor human-centric spaces consist of tabletop surfaces with hand-scale objects. In total, we collected 553~images for the 31~scenes, 4 of which were set aside for validation. 

\parahead{Synthetic Objects}
To render realistic looking objects in the above real scenes, we picked hand-scale objects from ShapeNetCore~\cite{chang2015shapenet}, manually removing any that did not look real or had topology problems. In total, we picked 6 object categories---\emph{bottle, bowl, camera, can, laptop}, and \emph{mug}. We also created a \emph{distractor} category consisting of object instances from categories not listed above such as \emph{monitor, phone}, and \emph{guitar}. This improves robustness when making predictions for our primary categories even if other objects are present in the scene. Our curated version of ShapeNetCore consists of 1085~individual object instances of which we set aside 184~instances for validation.

\parahead{Context-Aware Compositing}
To improve realism, we composite virtual objects in a context-aware manner \ie, we place then where they would naturally occur (\eg, on supporting surfaces) with plausible lighting. We use a plane detection algorithm~\cite{feng2014fast} to obtain pixel-level plane segmentation in real images. Subsequently, we sample random locations and orientations on the segmented plane where synthetic objects could be placed. We then place several virtual light sources to mimic real indoor lighting conditions. Finally, we combine the rendered and real images to produce a realistic composite with perfect ground truth NOCS maps, masks, and class labels.

In total, we render \textbf{300K composited images}, of which 25K are set aside for validation. To our knowledge, this the largest dataset for category-level 6D pose and size estimation. Our mixed reality compositing technique was implemented using the Unity game engine~\cite{unity} with custom plugins for plane detection and point sampling (all of which will be publicly released). The images generated using our method look plausible and realistic resulting in improved generalization compared to using non-context aware data.

\subsection{Real-World Data}
To further improve and validate our algorithm's real-world performance under challenging clutter and lighting conditions, we captured two real-world datasets: (1)~a real-world \emph{training dataset} that supplements the mixed reality data we generated earlier, (2)~a real-world \emph{testing dataset} to evaluate the performance of 6D pose and size estimation.
We developed a semi-automatic method to annotate ground truth object pose and size. Figure~\ref{fig:dataset} shows examples of our real-world data.

We captured \textbf{8K~RGB-D frames} (4300 for training, 950 for validation and 2750 for testing) of 18~different real scenes (7 for training, 5 for validation, and 6 for testing) using a Structure Sensor~\cite{ssensor}. For each of the training and testing subsets, we used 6~categories and 3~unique instances per category. For the validation set we use 6~categories with 1~unique instance per category. We place more than 5~object instances in each scene to simulate real-world clutter. For each instance, we obtained a clean and accurate 3D mesh using an RGB-D reconstruction algorithm that we developed for this purpose. In total, our combined datasets contain \textbf{18~different real scenes}, \textbf{42~unique object instances} spanning 6 categories making it the most comprehensive dataset for category-level 6D pose and size estimation.

\section{Method}
\label{sec:method}
Figure~\ref{fig:pipeline} shows our method for 6D pose and size estimation of multiple previously unseen objects from an \mbox{RGB-D} image. A CNN predicts class labels, masks, and NOCS maps of objects. We then use the NOCS map and the depth map to estimate the metric 6D pose and size of objects.

\subsection{NOCS Map Prediction CNN}
The goal of our CNN is to estimate class labels, instance masks, and NOCS maps of objects based purely on RGB images. We build upon the region-based Mask R-CNN framework~\cite{he2017mask} since it has demonstrated state-of-the-art performance on 2D object detection and instance segmentation tasks, is modular and flexible, fast, and can easily be augmented to predict NOCS maps as described below.

\subsubsection{NOCS Map Head}
Mask R-CNN builds upon the Faster R-CNN architecture~\cite{ren2015faster} and consists of two modules---a module to propose regions potentially containing objects, and a detector to detect and classify objects within regions. Additionally, it also predicts the instance masks of objects within the regions.
\begin{figure}[th!]
\centering
\vspace{-3mm}
  \includegraphics[width=\columnwidth]{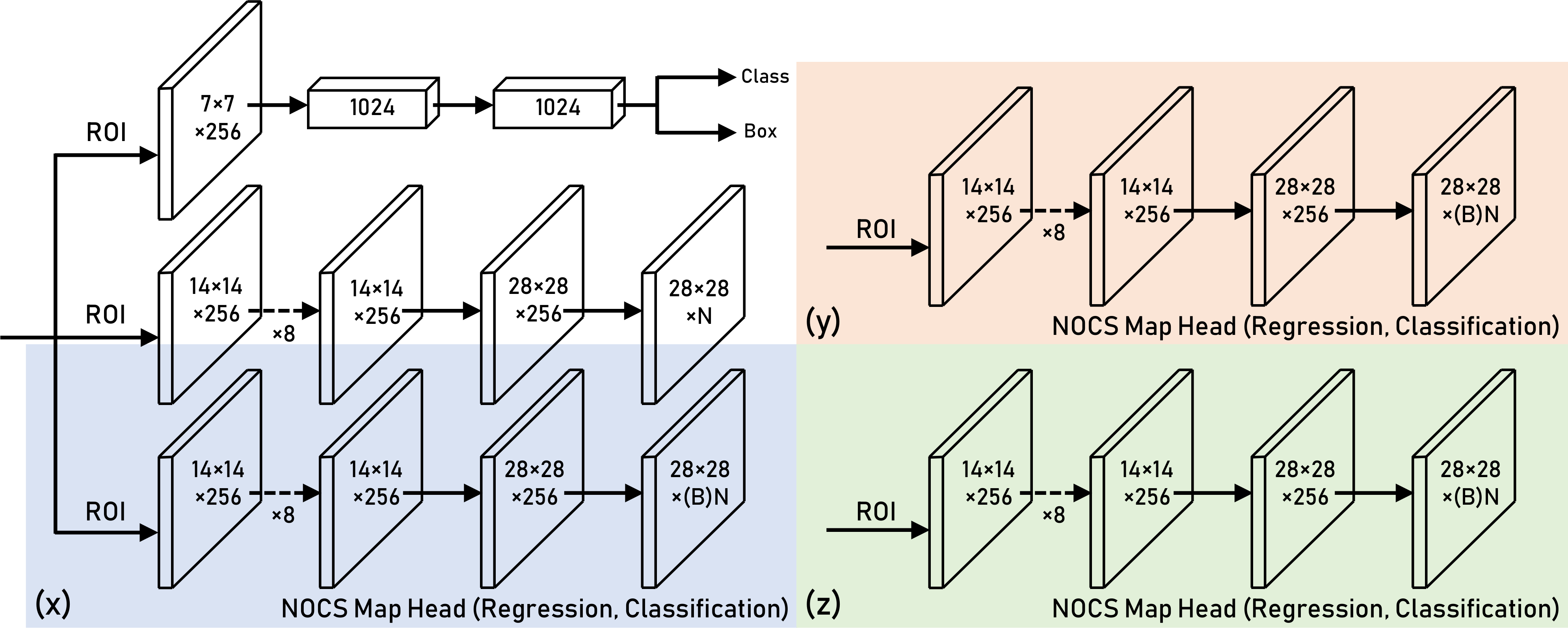} 
  \caption{NOCS map head architecture. We add three additional heads to the Mask R-CNN architecture to predict the $x, y, z$ coordinates of the NOCS map (colored boxes). These heads can either be used for direct pixel regression or classification (best). We use ReLU activation and 3$\times$3 convolutions.}
  \label{fig:nochead}
   \vspace{-3mm}
\end{figure}

Our main contribution is the addition of 3 head architectures to Mask R-CNN for predicting the $x, y, z$ components of the NOCS maps (see Figure~\ref{fig:nochead}). For each proposed region of interest (ROI), the output of a head is of size 28$\times$28$\times$N, where $N$ is the number of categories and each category containing the $x$ (or $y, z$) coordinates for all detected objects in that category. Similar to the mask head, we use the object category prior to look up the corresponding prediction channel during testing. During training, only the NOCS map component from the ground truth object category is used in the loss function. We use a ResNet50~\cite{he2016deep} backbone together with Feature Pyramid Network (FPN).

\parahead{Regression vs.\ Classification}
To predict the NOCS map, we can either regress each pixel value or treat it as a classification problem by discretizing the pixel values (denoted by (B) in Figure~\ref{fig:nochead}). Direct regression is presumably a harder task with the potential to introduce instability during training. Similarly, pixel classification with large number of classes (\eg, $B=128, 256$) could introduce more parameters making training even more challenging than direct regression. Our experiments revealed that pixel classification with $B=32$ performed better than direct regression.

\parahead{Loss Function}
The class, box, and mask heads of our network use the same loss functions as described in \cite{he2017mask}. For the NOCS map heads, we use two loss functions: a standard softmax loss function for classification, and the following soft $L^1$ loss function for regression which makes learning more robust.
%
\newcommand{\YGT}{\mathbf{y}}
\newcommand{\YGTS}{\mathbf{\tilde{y}}}
\newcommand{\YPred}{\mathbf{y}^\ast}
\begin{align}
L(\YGT, \YPred) = \frac{1}{n}
\begin{cases} 
      5 \, (\YGT - \YPred)^2, & |\YGT - \YPred| \leq 0.1 \\
      |\YGT - \YPred| - 0.05, & |\YGT - \YPred| > 0.1
\end{cases},\nonumber \\
\forall \YGT \in N, \YPred \in N_p, \nonumber
\end{align}
where $\YGT \in \mathcal{R}^3$ is the ground truth NOCS map pixel value, $\YPred$ is the predicted NOCS map pixel value, $n$ is the number of mask pixels inside the ROI, $I$ and $I_p$ are the ground truth and predicted NOCS maps.

\parahead{Object Symmetry}
Many common household objects (\eg, bottle) exihibit symmetry about an axis. Our NOCS representation does not take symmetries into account which resulted in large errors for some object classes. To mitigate this issue, we introduce a variant of our loss function that takes symmetries into account. For each category in our training data, we define an axis of symmetry. Pre-defined rotations about this axis result in NOCS maps that produce identical loss function values. For instance, a cuboid with a square top has a vertical symmetry axis. Rotation by angles, $\theta = \{0^{\circ}, 90^{\circ}, 180^{\circ}, 270^{\circ}\}$ on this axis leads to identical NOCS maps and therefore have the same loss.
For non-symmetric objects, $\theta = 0^{\circ}$ is unique. We found that a $|\theta| \le 6$ is enough to handle most symmetric categories. We generate ground truth NOCS maps, $\{\YGTS_1, \ldots, \YGTS_{|\theta|}\}$, that are rotated $|\theta|$ times along the symmetry axis. We then define our symmetric loss function, $L_s$ as
$
L_s = \min_{i = 1, \ldots,|\theta|} L\left(\YGTS_i, \YPred \right),\nonumber
$
where $\YPred$ denotes the predicted NOCS map pixel $(x, y, z)$.

\parahead{Training Protocol}
We initialize the ResNet50 backbone, RPN and FPN with the weights trained on 2D instance segmentation task on the COCO dataset\cite{lin2014microsoft}. For all heads, we use the initialization technique proposed in \cite{he2015delving}. We use a batch size of 2, initial learning rate of 0.001, and an SGD optimizer with a momentum of 0.9 and a 1$\times$10$^{-4}$ weight decay. In the first stage of training, we freeze the ResNet50 weights and only train the layers in the heads, the RPN and FPN for 10K iterations. In the second stage, we freeze ResNet50 layers below level 4 and train for 3K iterations. In the final stage, we freeze ResNet50 layers below level 3 for another 70K iterations. When switching to each stage, we decrease the learning rate by a factor of 10.

\subsection{6D Pose and Size Estimation}
Our goal is to estimate the full metric 6D pose and dimensions of detected objects by using the NOCS map and input depth map. To this end, we use the RGB-D camera intrinsics and extrinsics to align the depth image to color image. We then apply the predicted object mask to obtain a 3D point cloud $P_m$ of the detected object. We also use the NOCS map to obtain a 3D representation of $P_n$. We then estimate the scales, rotation, and translation that transforms the $P_n$ to $P_m$. We use the Umeyama algorithm~\cite{umeyama1991least} for this 7 dimensional rigid transformation estimation problem, and RANSAC~\cite{fischler1987random} for outlier removal. Please see the supplementary materials for qualitative results.

\section{Experiments and Results}
\label{sec:results}
\parahead{Metrics}
We report results on both 3D object detection, and 6D pose estimation metrics. To evaluate 3D detection and object dimension estimation, we use the intersection over union (IoU) metric with a threshold of 50\%~\cite{geiger2013vision}. For 6D pose estimation, we report the average precision of object instances for which the error is less than $m$~cm for translation and $n^{\circ}$ for rotation similar to \cite{shotton2013scene,li2018deepim}. We decouple object detection from 6D pose evaluation since it gives a clearer picture of performance. We set a detection threshold of 10\% bounding box overlap between prediction and ground truth to ensure that most objects are included in the evaluation. For symmetric object categories (\emph{bottle, bowl}, and \emph{can}), we allow the predicted 3D bounding box to freely rotate around the object's vertical axis with no penalty. We perform special processing for the \emph{mug} category by making it symmetric when the handle is not visible since it is hard to judge its pose in such cases, even for humans. We use \cite{yi2016scalable} to detect handle visibility for CAMERA data and manually annotate for real data.

\parahead{Baselines}
Since we know of no other methods for category-level 6D pose and size estimation, we built our own baseline to help compare performance. It consists of the Mask R-CNN network trained on the same data but without the NOCS map heads. We use the predicted instance mask to obtain a 3D point cloud of the object from the depth map.
We align (using ICP~\cite{besl1992method}) the masked point cloud to one randomly chosen model from the corresponding category. For instance-level 6D pose estimation, we present results that can readily be compared with \cite{xiang2017posecnn}.

\parahead{Evaluation Data}
All our experiments use one or both of these evaluation datasets: (1)~the CAMERA validation dataset (CAMERA25), and (2)~a 2.75K real dataset (REAL275) with ground truth annotations. Since real data is limited, this allows us to investigate performance without entangling pose estimation and domain generalization.

\subsection{Category-Level 6D Pose and Size Estimation}
\label{sec:main_result}
\begin{figure}[th!]
\centering
  \includegraphics[width=\linewidth]{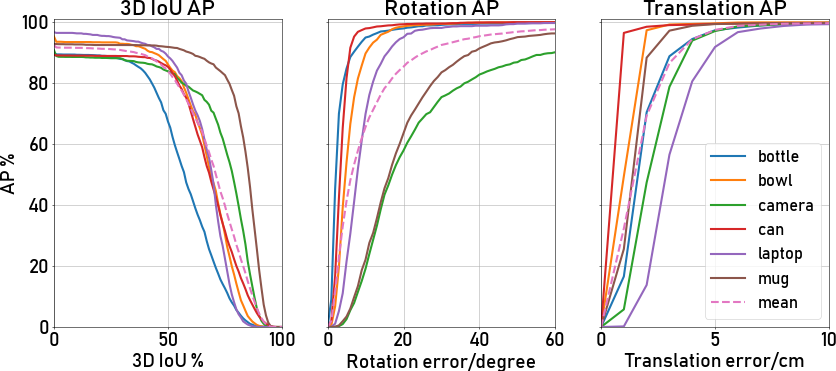}
  \caption{3D detection and 6D pose estimation results on CAMERA25 when our network is trained on CAMERA*.}
  \label{fig:camera_val_result}
  \vspace{-3mm}
\end{figure}

\parahead{Test on CAMERA25}
We report category-level results for our method with the CNN trained only on the 275K CAMERA training set (CAMERA*). We test performance on CAMERA25 which is composed of objects and backgrounds completely unseen during training. We achieve a mean average precision (mAP) of \textbf{83.9\%} for 3D IoU at 50\% and an mAP of \textbf{40.9\%} for the ($5^{\circ},\, 5$~cm) metric. ($5^{\circ},\, 5$~cm) is a strict metric for estimating 6D pose even for known instances~\cite{xiang2017posecnn,brachmann2016uncertainty,rad2017bb8}. See Figure~\ref{fig:camera_val_result} for more details.
%

\begin{figure*}[th!]
\centering
\vspace{-7mm}
  \includegraphics[width=\linewidth]{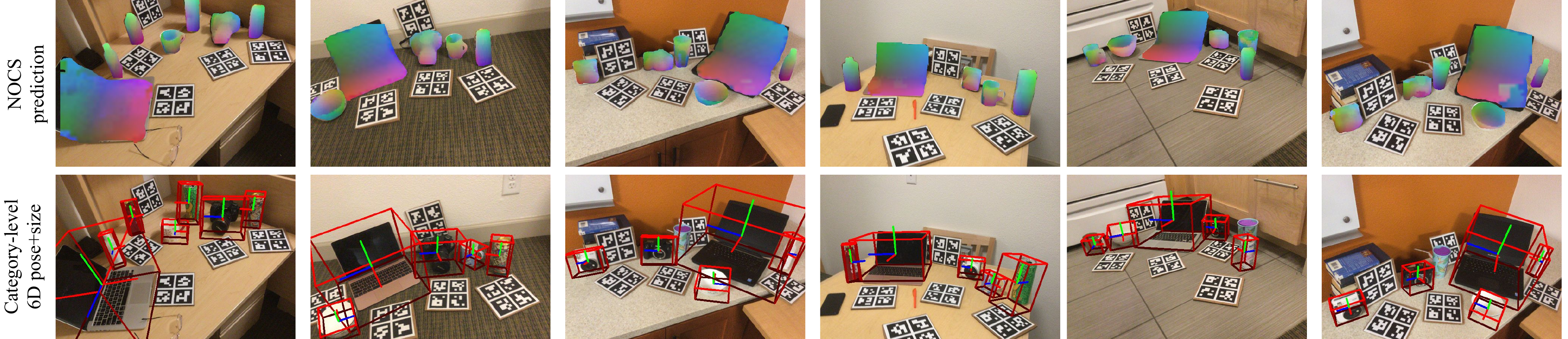}
  \vspace{-4mm}
  \caption{Qualitative result on REAL275 test set. Top row shows the predicted NOCS maps color coded. Bottom row shows the quality of 6D pose (axis) and size estimation (red tight bounding box).}
  \label{fig:result}
  \vspace{-3mm}
\end{figure*}

\parahead{Test on REAL275}
We then train our network on a combination of CAMERA*, the real-world dataset (REAL*), with weak supervision from COCO~\cite{lin2014microsoft}, and evaluate it on the real-world test set. 
Since COCO does not have ground truth NOCS maps, we do not use NOCS loss during training. We use 20K COCO images that contain instances in our categories. To balance between these datasets, for each minibatch we select images from the three data sources, with a probability of 60\% for CAMERA*, 20\% for COCO, and 20\% for REAL*. This network is the best performing model which we use to produce all visual results (Figure \ref{fig:result}).

\begin{figure}[th!]
\centering
  \includegraphics[width=\linewidth]{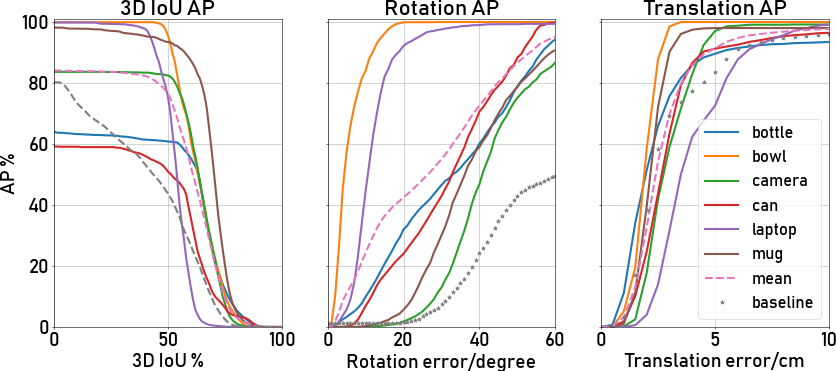} 
  \caption{Result on REAL275 test set, average precision (AP) vs.\ different thresholds on 3D IoU, rotation error, and translation error.}
  \label{fig:real_test_result}
  \vspace{-3mm}
\end{figure}

In the real test set, we achieved an mAP of \textbf{76.4\%} for 3D IoU at 50\%, an mAP \textbf{10.2\%} for the ($5^{\circ},\, 5$~cm) metric, and an mAP of \textbf{23.1\%} for ($10^{\circ},\, 5$~cm) metric. 
In comparison, the baseline algorithm (Mask RCNN + ICP alignment) achieves an mAP of \textbf{43.8\%} for 3D IoU at 50\%,and an mAP of \textbf{0.8\%} for both ($5^{\circ},\, 5$~cm) and ($10^{\circ},\, 5$~cm) metric, which is significantly lower than our algorithm's performance.  Figures~\ref{fig:real_test_result} shows more detailed analysis and comparison.  
This experiment demonstrates that by learning to predict the dense NOCS map, our algorithm is able to provide additional detailed information about the object's shape, parts and visibility, which are all critical for 
correct estimation the object's 6D pose and sizes. 

\subsection{Ablation Studies}
\parahead{CAMERA Approach}
\label{sec:data_ablation}
To evaluate our CAMERA data generation approach, we conducted an experiment with our network trained on different training data combinations. For this experiment, we set the network architecture to regress the NOCS maps. Table~\ref{table:data_ablation} shows the performance of our network on the REAL275 test set.

We also created a variant of CAMERA* where the images are composited in a \textbf{non-context aware} manner (denoted by \textbf{B} in Table~\ref{table:data_ablation}). As shown in the table, using only CAMERA* results in poor performance due to domain gap. We see progressive improvements on adding COCO and REAL*. Training only on REAL*, or REAL* and COCO tend to overfit to the training data due to small dataset size. Training on CAMERA* with COCO and REAL* lead to the best results. Furthermore, we see that non-context aware data results in worse performance than context-aware data indicating that our CAMERA approach is useful.

\begin{table}[h]
  \centering
  \footnotesize
  \setlength{\tabcolsep}{4.0 pt}
  \begin{tabular}{c|c|c|c|c|c|c|c}
    \toprule
    \multicolumn{3}{c|}{\textbf{Data}} & \multicolumn{5}{c}{\textbf{mAP}}\\ \midrule
    \multirow{2}{*}{CAMERA*} & \multirow{2}{*}{COCO} & \multirow{2}{*}{REAL*} & \multirow{2}{*}{$\text{3D}_{25}$}  &  \multirow{2}{*}{$\text{3D}_{50}$} & {5 $^{\circ}$} &  {10$^{\circ}$} & {10$^{\circ}$}\\ 
       &  &  &  &  &  5 cm & 5 cm  & 10cm\\ \hline
     \textbf{C} &  &  & 51.7 & 36.7 & 3.4 & 20.4 & 21.7\\ 
     \textbf{C} & \checkmark & &57.6&41.0&3.3&17.0&17.1\\ 
     &  & \checkmark & 61.9 & 47.5 & 6.5 & 18.5 & 18.6\\
     & \checkmark & \checkmark &71.0&53.0&7.6&16.3&16.6\\
     \textbf{C} &  & \checkmark &79.2&69.7&6.9&20.0&21.2\\ 
     \textbf{C} & \checkmark & \checkmark &  \textbf{79.6} & \textbf{72.4} &  \textbf{8.1} & \textbf{23.4} &\textbf {23.7}\\
      \midrule
     \textbf{B} &  &  & 42.6 & 36.5 & 0.7 & 14.1 & 14.2\\ 
     \textbf{B} & \checkmark & \checkmark & 79.1 & 71.7 & 7.9 & 19.3 & {19.4}\\\bottomrule
    \end{tabular}
  \caption{Validating CAMERA approach. \textbf{C} represents the unmodified CAMERA* data while \textbf{B} denotes a non-context aware version of CAMERA*. We report AP for 5 different metrics, where $\text{3D}_{25}$ and $\text{3D}_{25}$ represent 3D IoU at 25\% and 50\%, respectively.}
  \label{table:data_ablation}
   \vspace{-3mm}
\end{table}

\parahead{Classification vs.\ Regression}
On both CAMERA25 and REAL275, pixel classification is consistently better than regression. Using 32 bins is best for pose estimation while 128 bins is better on detections (see Table~\ref{table:network_ablation}).

\parahead{Symmetry Loss}
This loss is critical for many everyday symmetric object categories. To study the effect of symmetry loss, we conduct ablation experiments on the regression network on both CAMERA25 and REAL275 set. Table~\ref{table:network_ablation} shows that the pose accuracy degrades significantly, particularly for 6D pose, if the symmetry loss is not used.
\begin{table}[h]
\footnotesize
  \centering
  \setlength{\tabcolsep}{4.0 pt}
  \begin{tabular}{c|c|c|c|c|c|c}
    \toprule
    \multirow{3}{*}{\textbf{Data}} & \multirow{3}{*}{\textbf{Network}} & \multicolumn{5}{c}{\textbf{mAP}}\\ \cline{3-7}
     &   & \multirow{2}{*}{$\text{3D}_{25}$}  &  \multirow{2}{*}{$\text{3D}_{50}$} & {5 $^{\circ}$} &  {10$^{\circ}$} & {10$^{\circ}$}\\ 
       &  &  &  &    5 cm & 5 cm  & 10cm\\ \midrule
     \multirow{5}{*}{\textbf{CAMERA25}} & {Reg.}  & 89.3 & 80.9 & 29.2 & 53.7 & {54.5}\\
      & {Reg. w/o\ Sym.}  & 86.6 & 79.9 & 14.7 & 38.5 & {40.0}\\
      & {32 bins}  & 91.1 & 83.9 & \textbf{40.9} & \textbf{64.6} & \textbf{65.1}\\
      & {128 bins}  & \textbf{91.4} & \textbf{85.3} & 38.8& 61.7 & {62.2}\\\midrule
     \multirow{5}{*}{\textbf{REAL275}} & {Reg.}  & 79.6 & 72.4 & 8.1 & 23.4 & {23.1}\\
      & {Reg. w/o Sym.} & 82.7& 73.8 & 1.3 & 9.1  & 9.3  \\
      & {32 bins}    & 84.8&  78.0& \textbf{10.0}& {25.2} & {25.8}\\
      & {128 bins} & \textbf{84.9} & \textbf{80.5} & 9.5 &\textbf{ 26.7}  & \textbf{26.7} \\\bottomrule
    \end{tabular}
  \caption{Network architectures and losses. Reg. represents regression network trained with soft $L^1$ loss; 32 bins and 128 bins represent classification networks with the corresponding numbers of bins, respectively.}
    \label{table:network_ablation}
     \vspace{-3mm}
\end{table}

\subsection{Instance-level 6D Pose Estimation}
We also evaluate our method on instance-level 6D pose estimation task on OccludedLINEMOD~\cite{LINEMOD} and compare with PoseCNN~\cite{xiang2017posecnn}.
%
The OccludedLINEMOD dataset has 9 object instances 
and provides a CAD model for each instance. It has 1214 images with annotated ground truth 6D pose. 
We follow the protocols from \cite{tekin2017,kehl2017ssd} and randomly select $15\%$ of the dataset as training images. We then generate 15000 synthetic images using the technique described in Section~\ref{sec:dataset}.
\begin{figure}[th!]
\centering
  \includegraphics[width=\linewidth]{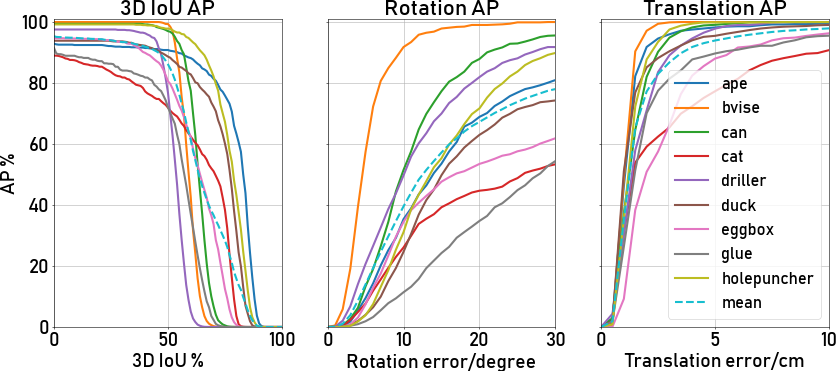}
  \caption{Result on OcculudedLINEMOD. Here we show the average precision (AP) vs. different thresholds on 3D IoU, rotation error, and translation error.}
  \label{fig:linemod_result}
  \vspace{-3mm}
\end{figure}

Using 32-bin classification network, we achieve a detection rate of 94.7\%, an mAP of \textbf{88.4\%} for 3D IoU at 50\%, an mAP \textbf{13.9\%} for the ($5^{\circ},\, 5$~cm) metric, and an mAP of \textbf{33.5\%} for ($10^{\circ},\, 5$~cm) metric. This is substantially higher than PoseCNN~\cite{xiang2017posecnn} which only achieves an mAP of \textbf{1.7\%} without iterative pose refinement (reported in \cite{li2018deepim}). Figure~\ref{fig:linemod_result} provide a more detailed analysis. This experiment demonstrates that while our approach designed for category-level pose estimation, it can also achieve state-of-the-art performance on standard 6D pose estimation benchmarks. 

With the 2D projection metric, which measures the average pixel distance between ground truth and estimated object poses, we achieve \textbf{30.2\%}~mAP on 2D projection at 5 pixel. Our method significantly outperforms PoseCNN~\cite{xiang2017posecnn} by a large margin, which reported 17.2\%~mAP on 2D projection at 5 pixel in \cite{li2018deepim}. Please see the supplementary document for detailed comparison.
%

\parahead{Limitations and Future Work}
To our knowledge, ours is the first approach to solve the category-level 6D pose and size estimation problem. 
There are still many open issues that need to be addressed. First, in our approach, the pose estimation is conditioned on the region proposals and category prediction which could be incorrect and negatively affect the results. Second, our approach rely on the depth image to lift NOCS prediction to real-world coordinates. Future work should investigate estimating 6D pose and size directly from RGB images.

\section{Conclusion}
We presented a method for category-level 6D pose and size estimation of previously unseen object instances. We presented a new normalized object coordinate space (NOCS) that allows us to define a shared space with consistent object scaling and orientation. We propose a CNN that predicts NOCS maps that can be used with the depth map to estimate the full metric 6D pose and size of unseen objects using a pose fitting method. Our approach has important applications in areas like augmented reality, robotics, and 3D scene understanding.
{
  \small
\parahead{Acknowledgements}
This research was supported by a grant from Toyota-Stanford Center for AI Research, NSF grant IIS-1763268, a gift from Google, and a Vannevar Bush Faculty Fellowship. We thank Xin Wang, Shengjun Qin, Anastasia Dubrovina, Davis Rempe, Li Yi, and Vignesh Ganapathi-Subramanian.
}

\appendix
\section{Implementation and Computation Times}
Our network is implemented on Python 3, Keras and Tensorflow. The code is based on MatterPort's Mask RCNN implementation\cite{matterport_maskrcnn_2017}. The network uses Feature Pyramid Network (FPN)\cite{lin2017feature} and a ResNet50 backbone\cite{he2016deep}.

Our network takes images with a resolution of 640$\times$360 as input. We achieve an interactive rate of around 4 fps on an Intel Xeon Gold 5122 CPU @ 3.60GHz desktop with a NVIDIA TITAN Xp. Our implementation takes an average time of 210 ms for neural network inference and 34 ms for pose alignment using Umeyama algorithm. 

\begin{figure*}[th]
\vspace{-7mm}
  \includegraphics[width=\linewidth]{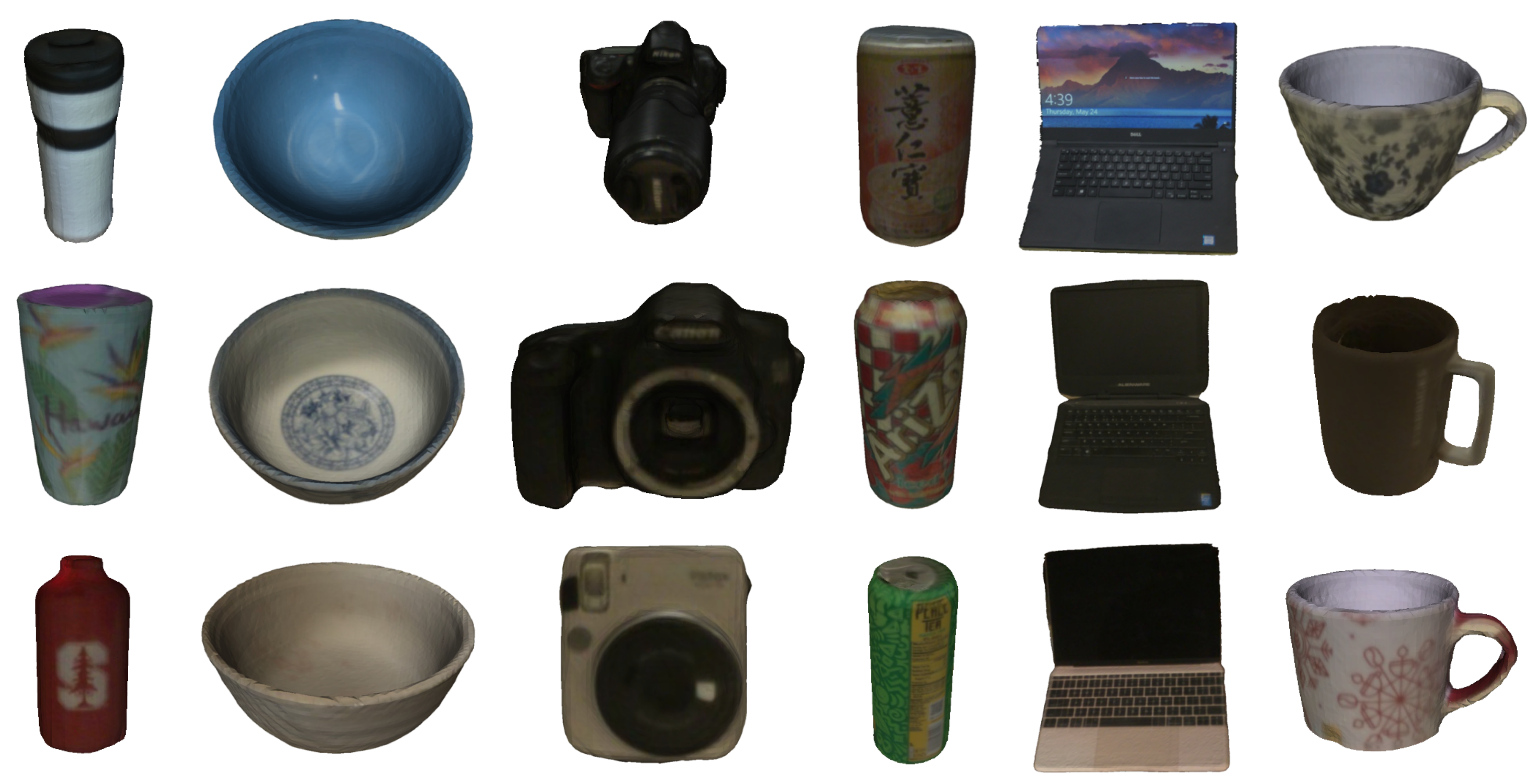} \vspace{-4mm}
  \caption{A subset of our 42 instances in our real dataset.  From the left to the right column, they are: bottle, bowl, camera, can, laptop and mug. The first row shows instances that belong to training set. The second and third rows show instances that belong to test set.}
  \label{fig:instances}
  \vspace{-3mm}
\end{figure*}

\section{Scanned Real Instances}
Our real dataset contains 6 object categories and 42 real scanned unique instances. For each category, we collect 7 instances with 4 for training and validation and the rest 3 for test. Figure \ref{fig:instances} show a subset of our instances where one can see a large intra-category shape variance in the dataset. The first row are instances used in training. The second and third rows are held-out instances for testing.  

\section{Result Visualization}
\label{supp_visual}
\begin{figure*}[th!]
\centering
\vspace{0mm}
  \includegraphics[width=\linewidth]{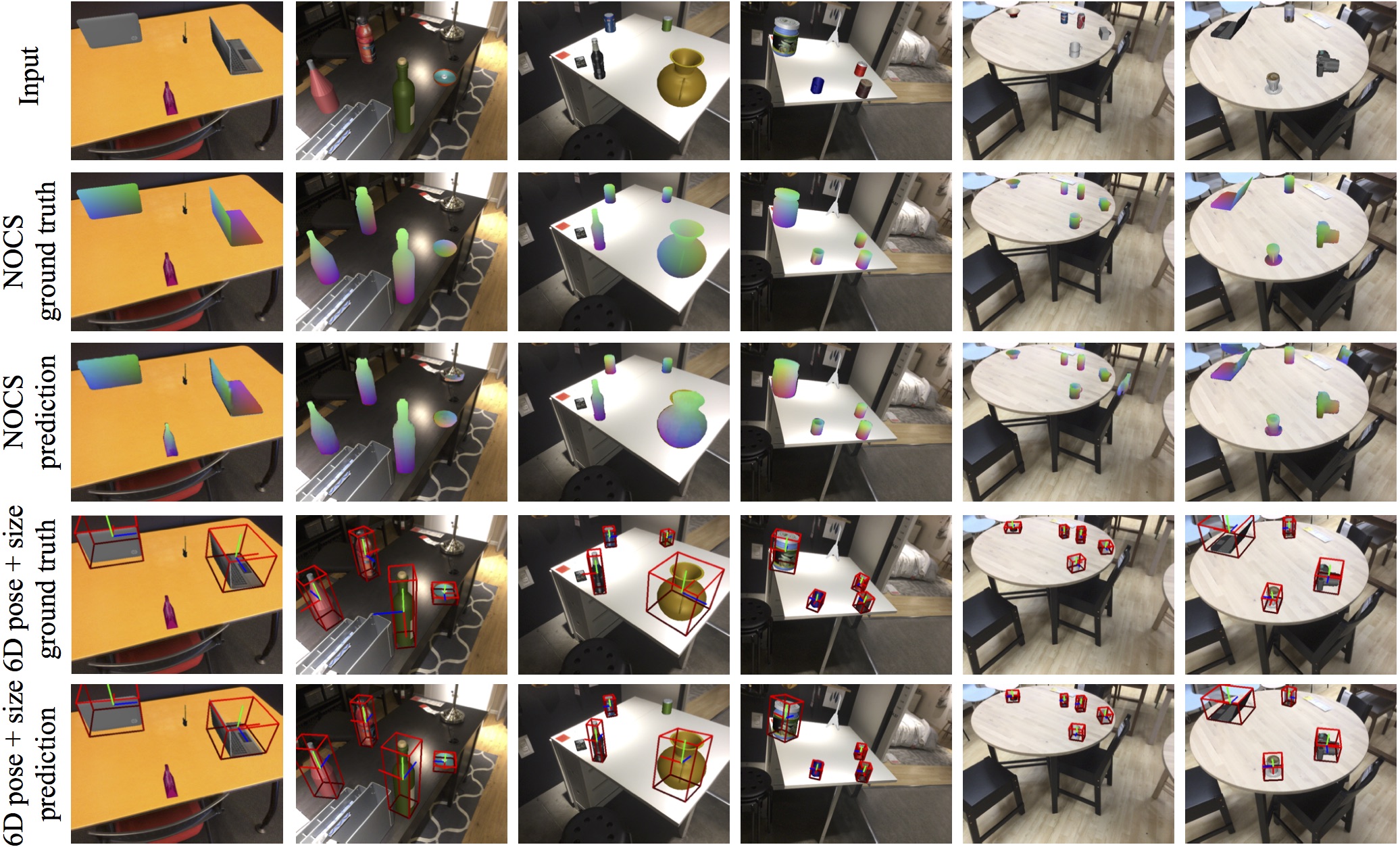} \vspace{-4mm}
  \caption{Qualitative result on CAMERA25 validationset set. The top row shows the input rgb image. The second and the third rows show the color-coded ground truth and predicted NOCS maps. The fourth and the fifth rows show the ground truth and predicted 6D pose (axis) and size estimation (red tight bounding box).}
  \label{fig:supp_val_result}
  \vspace{-1mm}
\end{figure*}

Here we provide more visual results of the 6D pose and size estimation. Due to sufficient training data, our method achieves very promising performance on CAMERA25 validation set as shown in Figure\ref{fig:supp_val_result}. On REAL275 test set, we still observe decent performance even though the amount of real training data is small. We observe several failure modes on real data, including missing detection, wrong classification, and inconsistency in predicted coordinate maps.

\begin{figure*}[th!]
\centering
\vspace{-1mm}
  \includegraphics[width=\linewidth]{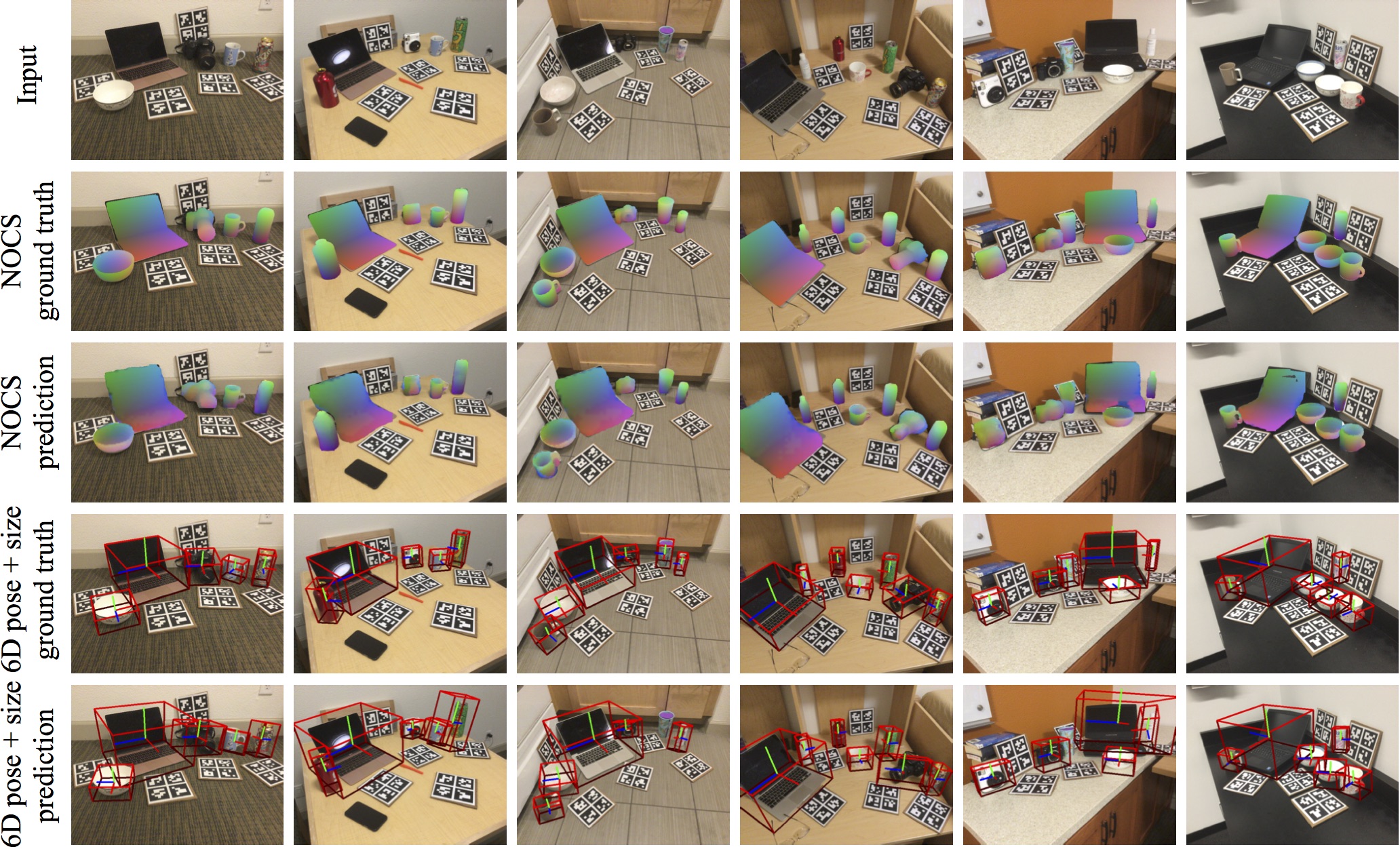} \vspace{-4mm}
  \caption{Qualitative result on REAL275 test set. }
  \label{fig:supp_real_result}
  \vspace{-3mm}
\end{figure*}

\section{Comparisons on the OccludedLINEMOD Dataset}
\begin{figure*}[th]
  \includegraphics[width=\linewidth]{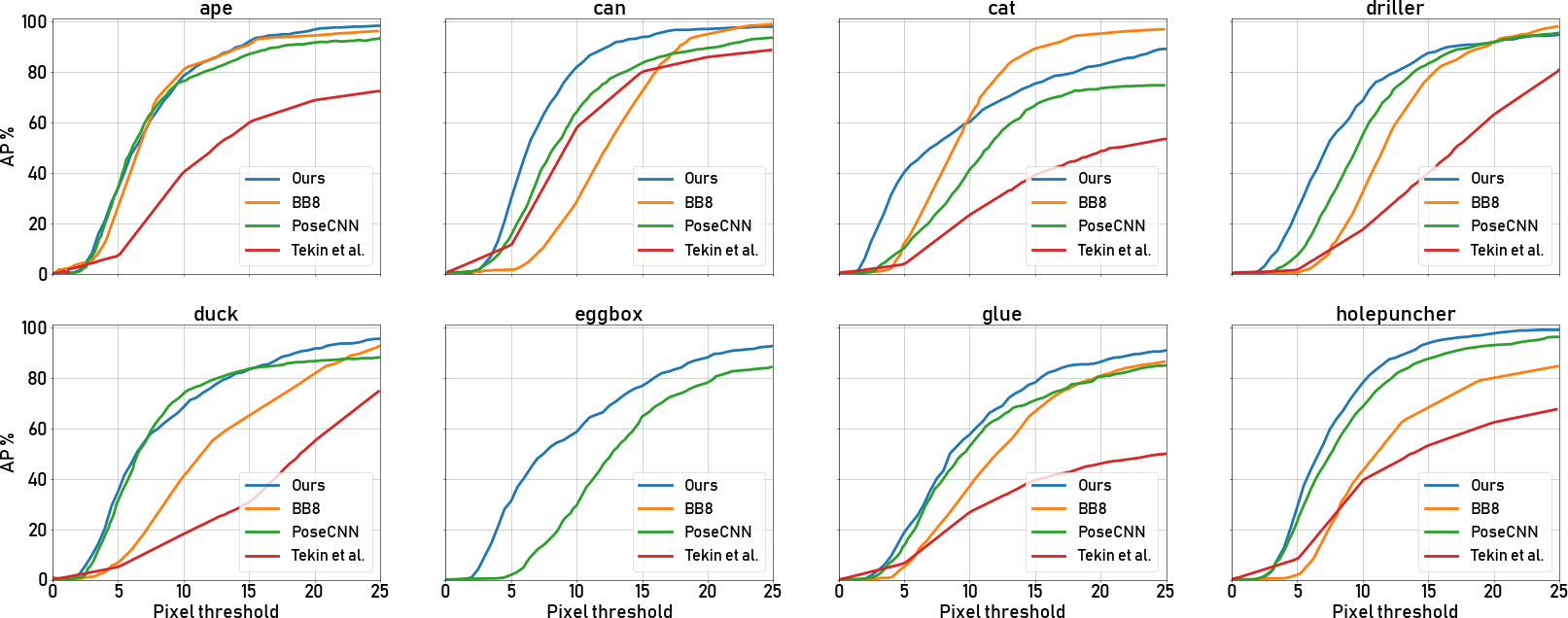} 
  \caption{Comparison with state-of-the-art RGB or RGB-D based methods on the OccludedLINEMOD dataset\cite{brachmann2016uncertainty}. }
  \label{fig:supp_linemod_result}
\end{figure*}

The comparison between our method and other existing methods with 2D projection metric on OccludedLINEMOD dataset\cite{brachmann2016uncertainty} is shown in Figure~\ref{fig:supp_linemod_result}.

{\small
\bibliographystyle{ieee}
\bibliography{NOCS_arXiv}
}

\end{document}